\documentclass{bmvc2k}
\usepackage{multirow}
\usepackage{hhline}
\usepackage{booktabs}
\usepackage{amssymb}

\title{Dynamic Feature Alignment for \\ Semi-supervised Domain Adaptation}

\addauthor{Yu Zhang}{y.zhang@uky.edu}{1}
\addauthor{Gongbo Liang}{gongbo.liang@eku.edu}{2}
\addauthor{Nathan Jacobs}{jacobs@cs.uky.edu}{1}

\addinstitution{
 Department of Computer Science\\
 University of Kentucky\\
 Lexington, KY, USA
}
\addinstitution{
 Department of Computer Science\\
 Eastern Kentucky University\\
 Richmond, KY, USA
}

\runninghead{Zhang, Liang, Jacobs}{Dynamic Feature Alignment for SSDA}

\begin{document}

\maketitle

\begin{abstract}

Most research on domain adaptation has focused on the purely unsupervised setting, where no labeled examples in the target domain are available. However, in many real-world scenarios, a small amount of labeled target data is available and can be used to improve adaptation. We address this semi-supervised setting and propose to use dynamic feature alignment to address both inter- and intra-domain discrepancy. Unlike previous approaches, which attempt to align source and target features within a mini-batch, we propose to align the target features to a set of dynamically updated class prototypes, which we use both for minimizing divergence and pseudo-labeling. By updating based on class prototypes, we avoid problems that arise in previous approaches due to class imbalances. Our approach, which doesn't require extensive tuning or adversarial training, significantly improves the state of the art for semi-supervised domain adaptation. We provide a quantitative evaluation on two standard datasets, DomainNet and Office-Home, and performance analysis. Project Page: \href{https://mvrl.github.io/DFA}{https://mvrl.github.io/DFA}

\end{abstract}

\section{Introduction}
\label{sec:intro}

Deep neural networks have achieved impressive performance on a wide range of tasks, including image classification, semantic segmentation, and object detection. However, models often generalize poorly to new domains, such as when a model trained on indoor imagery is used to interpret an outdoor image.
%
\textit{Domain Adaptation} (DA) methods aim to take a model trained on a label-rich source domain and make it generalize well to a label-scarce target domain. Recently, most studies on domain adaptation have focused on the \textit{Unsupervised Domain Adaptation} (UDA) setting, in which no labeled target data is available. However, in real-world scenarios there is often a small amount of labeled target data available: the \textit{Semi-Supervised Domain Adaptation} (SSDA) setting. Recent works~\cite{kim2020attract, saito2019semi} demonstrate that directly applying UDA methods on the semi-supervised setting can actually hurt performance. Therefore, finding a way to effectively use the small amount of labeled target imagery is an important problem. We propose a novel approach that is tailored to the SSDA setting.

In addition to poor generalization in terms of model performance, the intermediate features for source and target domain inputs often display a significant domain shift. This motivates approaches that use feature alignment strategies~\cite{ganin2015unsupervised, cao2018unsupervised, dong2020cscl, dong2020can, hoffman2018cycada, zhu2017unpaired, jiangbidirectional, saito2019semi} to minimize the distances between source and target distributions. These methods address the shift between source and unlabeled target samples but ignore the shift within the target domain brought by the small number of labeled target samples. A recent work APE~\cite{kim2020attract} addresses this issue as \textit{intra-domain discrepancy}, and proposes three schemes---\textit{attraction, perturbation, and exploration}, to alleviate this discrepancy. \textit{Attraction} is used to push the unlabeled features to the labeled feature distribution. \textit{Perturbation} aims to move both labeled and unlabeled target features to their intermediate regions to minimize the gap in between. \textit{Exploration} is complementary to the other two schemes by selectively aligning unlabeled target features to the source. 

Due to the imbalance between the large amount of labeled source data and the small amount of labeled target data as well as the class imbalance (e.g.\ the existence of long-tail classes), a random mini-batch of aligned features can not always represent the true distribution of the data. Therefore, the alignment of unlabeled features can be inaccurate. Moreover, errors can be accumulated when incorrectly predicted unlabeled samples are selected to be used for pseudo-label training during \textit{exploration}.

Considering the concerns mentioned above, we propose a novel \textit{Dynamic Feature Alignment} (DFA) framework for the SSDA problem. Instead of directly aligning the unaligned target features to the aligned features within a mini-batch, we propose to align the unlabeled target features to a set of dynamically updated class prototypes, which are stored in a dynamic memory bank $\mathcal{B}$. To utilize these prototypes for pseudo-labeling, we selectively collect unlabeled samples based on their distances to class prototypes and their prediction entropy. We evaluate our method on standard domain adaptation benchmarks, including DomainNet and Office-Home, and results show that our method achieves significant improvement over the state of the art in both the 1-shot and 3-shot settings.

\section{Related Work}


We introduce related work in unsupervised and semi-supervised domain adaptation and describe their relationship to our approach.

\subsection{Unsupervised Domain Adaptation}
UDA is a machine learning technique that trains a model on one or more source domains and attempts to make the model generalize well on a different but related target domain~\cite{ben2010theory, redko2019advances}. One of the key challenges of UDA is to mitigate the domain shift (or distribution shift) between the source and target domains. In general, three types of techniques can be used~\cite{wang2018deep, wilson2020survey}: (1) adversarial, (2) reconstruction based, and (3) divergence based.

The adversarial methods achieve domain adaptation by using adversarial training~\cite{dong2020can, hoffman2018cycada, zhu2017unpaired, jiangbidirectional, saito2019semi}. For instance, CoGAN~\cite{liu2016coupled} uses two generator/discriminator pairs for both the source and target domain, respectively, to generate synthetic data that is then used to train the target domain model. Domain-Adversarial Neural Networks (DANN)~\cite{ganin2016domain} promotes the emergence of features that are discriminative on the source domain and unable to discriminate between the domains. 

Reconstruction-based methods~\cite{ghifary2016deep, bousmalis2016domain, ghifary2015domain} use an auxiliary reconstruction task to create a representation that is shared by both domains. For instance, Deep Reconstruction Classification Network (DRCN)~\cite{ghifary2016deep} jointly learns a shared encoding representation from two simultaneously running tasks. Domain Seperation Networks (DSN)~\cite{bousmalis2016domain} propose a scale-invariant mean squared error reconstruction loss. Those learned representations preserve discriminability and encode useful information from the target domain.

While adversarial methods are often difficult to train and the reconstruction-based methods require heavy computational cost, divergence-based methods align the domain distributions by minimizing a divergence that measures the distance between the distributions during training with minimal extra cost. For instance, Maximum Mean Discrepancy (MMD)~\cite{gretton2006kernel} has been used in~\cite{rozantsev2018beyond} to align the features of two domains by using a two-branch neural network with unshared weights. Deep CORAL~\cite{sun2016deep} uses the Correlation Alignment (CORAL) ~\cite{sun2016return} as the divergence measurement and Contrastive Adaptation Network (CAN)~\cite{kang2019contrastive} measures the Contrastive Domain Discrepancy (CCD). Our proposed method DFA can be categorized as a divergence-based method.

\subsection{Semi-Supervised Domain Adaptation}
UDA approaches are effective at aligning source and target feature distributions without taking advantage of any labels from the target domain. In real-world scenarios, we usually have access to a small number of labeled target samples, which can be used to improve adaptation. This problem is addressed as semi-supervised domain adaptation (SSDA)~\cite{saito2019semi, jiangbidirectional, kim2020attract, motiian2017few, teshima2020few}. Conventional UDA methods can be applied to SSDA simply by combining the source data with labeled target data. However, due to the imbalance between the large amount of source data and the small amount of labeled target data as well as the class imbalance issue, this strategy may align target features misleadingly~\cite{ao2017fast, donahue2013semi, yao2015semi}.

To explore more effective solutions for the SSDA problem, BiAT~\cite{jiangbidirectional} proposes a bidirectional adversarial training method to effectively generate adversarial samples and bridge the domain shift. MME~\cite{saito2019semi} proposes a minimax entropy approach that adversarially optimizes an adaptive few-shot learning model. The key idea is to minimize the distance between the class prototypes and neighboring unlabeled target samples. The recent work APE~\cite{kim2020attract} extends MME~\cite{saito2019semi} by combining attraction, perturbation, and exploration strategies to bridge the intra-domain discrepancy. In contrast to previous works, we abandon the direct alignment between source and target features. Instead, our method is built upon a set of dynamically updated class prototypes.

\subsection{Memory Bank}
Memory banks have been applied in unsupervised learning and contrastive learning~\cite{wu2018unsupervised, he2020momentum, liu2018unsupervised} as a dictionary look-up to reduce the computational complexity of calculating distances or similarities between features. For instance, the memory bank in~\cite{wu2018unsupervised} is designed to compute the non-parametric softmax classifier more efficiently for large datasets. To learn discriminative features from unlabeled samples, it stores one instance per class and is updated with the new seen instances every iteration. The smoothness of the training was encouraged by adding a proximal regularization term, not a momentum update directly applied on the features. Another similar work is MoCo~\cite{he2020momentum}, which maintains a dynamic dictionary as a queue to replace old features with the current batch of features. Instead of applying the momentum on the representations, MoCo uses a momentum to keep the encoder slowly evolving during training. The dictionary size of MoCo can be very large, but it is not designed to be class-balanced during training when the size is small. Both~\cite{wu2018unsupervised, he2020momentum} are effective for contrastive learning but not ideal for SSDA when the goal is to learn stable, representative, and class-balanced prototypes for aligning features from different domains and pseudo-labeling. Our work aims to resolve this issue by designing a dynamic memory bank (introduced in Sec~\ref{sec: bank}) for better feature alignment.

\section{Method}

\begin{figure}[!tb]
    \centering
    \includegraphics[width=0.9\textwidth]{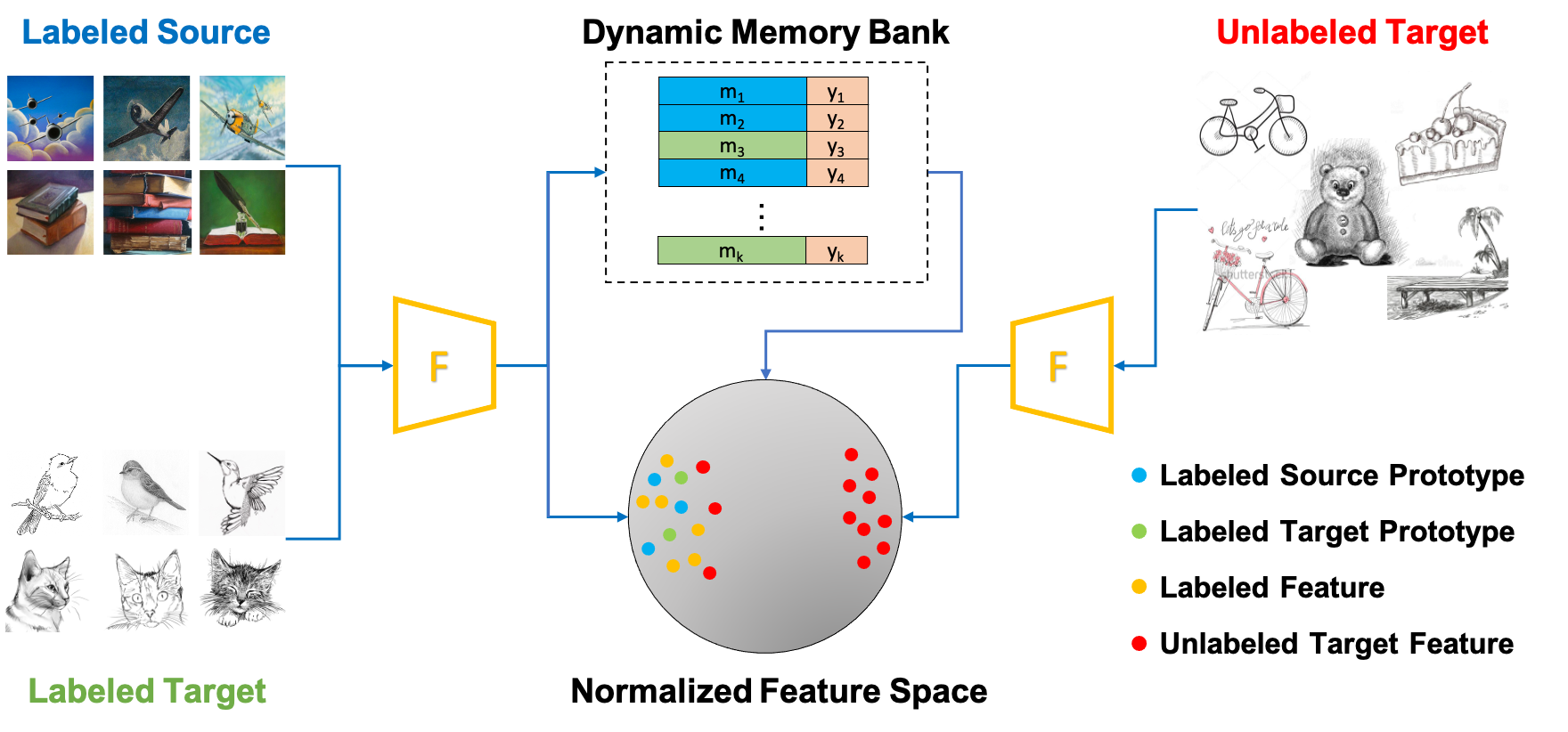}
    \caption{Framework of Dynamic Feature Alignment. Both labeled source and labeled target samples are passed into the feature extractor $F$. Normalized feature embeddings and labels are then stored in the dynamic memory bank as class prototypes for dynamic alignment and pseudo-labeling. The feature extractor for unlabeled target samples shares the same weights. }
    \label{fig:framework}
\end{figure}

We introduce \textit{Dynamic Feature Alignment} (DFA), a domain adaptation approach designed to work well in the semi-supervised setting. In the remainder of this section, we formalize the problem and describe the key components of our approach (see Fig~\ref{fig:framework} for an overview). 


\subsection{Problem Statement}

In \textit{SSDA}, we have the access to many labeled source domain samples $\mathcal{D}_{s} = {\{(x_i^s, y_i^s)\}_{i=1}^{n_s}}$ where $y_i^s \in \mathcal{Y} = \{1,...,Y\}$ with $n_s$ annotated pairs. In the target domain, we are also given a limited number of labeled target samples $\mathcal{D}_{l}=\left\{\left(x_{i}^{l}, {y_{i}}^{l}\right)\right\}_{i=1}^{n_{l}}$, as well as a relatively large number of unlabeled samples $\mathcal{D}_{u} = \left\{\left(x_{i}^{u} \right)\right\}_{i=1}^{n_{u}}$. Our goal is to train a robust model that can perform well on the target domain by making the most advantage of $\mathcal{D}_{s}$, $\mathcal{D}_{l}$, and $\mathcal{D}_{u}$.

\subsection{Supervised Classification in Normalized Feature Space}
\label{sec:supervised}



Our proposed DFA framework aligns the features of labeled examples from both domains using a supervised classification loss. A feature extractor $f(\cdot)$ is trained to extract the features from the input image ${x}$. The feature $m$, which is normalized to be unit length, is represented as:
\begin{equation}
    m = f(x).
\end{equation}
The classifier takes as inputs the normalized features $m$ and compares the cosine similarity between $m$ and the prototype weight vector $\mathbf{w}_k$ ($k$ = 1, ..., K) of class $k$. The similarities are scaled by a temperature $\tau=0.05$, which controls the concentration level of the distribution~\cite{hinton2015distilling, wang2017normface}. The probability of $m$ being categorized as class $k$ can be presented as:
\begin{equation}
    P(k|m) = \frac{exp(\mathbf{w}^{T}\cdot m / \tau)}{\sum_{j=1}^{K}exp(\mathbf{w}^{T}_{j}\cdot m / \tau)}.
\end{equation}
$P(k|m)$ is then passed to the classification loss~\cite{wu2018unsupervised, he2020momentum}. During training, we utilize cross-entropy as the supervised classification loss:
\begin{equation}
\mathcal{L}_{cls} = -\mathbb{E}_{(\mathbf{x}, y) \in \mathcal{D}_{s}
\cup\mathcal{D}_{l}} \sum_{k=1}^{K}\log P_k.
\end{equation}
When aligning the normalized features by training the network, the labeled target features $f(x^l)$ are closely aligned to the labeled source features $f(x^s)$~\cite{saito2019semi,kim2020attract}, and we aim to utilize both labeled source and target samples and reduce the domain shift.

\subsection{Dynamic Memory Bank}
\label{sec: bank}


We propose to maintain a dynamic memory bank $\mathcal{B}$ that stores representative features, which we call prototypes, for each class. If our feature extractor was fixed, we could simply extract all the features once and compute their averages. This isn't feasible since we are actively updating our feature extractor. We could recompute the prototypes periodically, but this would be computationally expensive. Instead we keep a weighted average of recently extracted features. A natural strategy for maintaining the memory bank would be to update the corresponding class prototype for each labeled sample in the current mini-batch. This could be done, for example, as an exponentially weighted moving average (EWMA). The downside of this approach is that the class prototype for common classes would update more frequently, which leads to difficulty in setting update weights. It also doesn't take into account the potential for a large domain shift to lead to less informative class prototypes, especially when many samples are misclassified.

To maintain our dynamic memory bank $\mathcal{B}$, we propose to use an intermediate memory bank $b$ to enable us to make consistent, class-balanced updates. For every minibatch, we update $b$ as follows: we check the network output $f(x_k)$ on the input image $x_k$, which could be from the source or target domain. If $x_k$ is correctly classified, then the corresponding vector in $b$ is replaced with $f(x_k)$. Therefore, $b$ always stores the feature of the most recent correctly classified image for each class. We use the intermediate memory bank $b$ to update $\mathcal{B}$, using an EWMA, as follows:
\begin{equation}
    \mathcal{B} = \gamma \cdot \mathcal{B}_{t-1} + (1-\gamma) \cdot b
\end{equation}
where $\gamma$ regulates the pace of the update: lower value results in a faster updating pace, and higher value leads to a slower update.

The dynamic memory bank $\mathcal{B}$ is updated based on labeled source features $f(x^s)$ and labeled target features $f(x^l)$. Therefore, each feature vector in $\mathcal{B}$ represents a class, and can be interpreted as a class prototype, so $\mathcal{B}$ stores the prototypes of all classes. In the following section, we describe how this dynamic memory bank can be applied to (1) align the unlabeled target features to class prototypes accurately and (2) selectively collect unlabeled samples for pseudo-label estimation.

 \begin{figure}[!tb]
    \centering
    \includegraphics[width=0.7\textwidth]{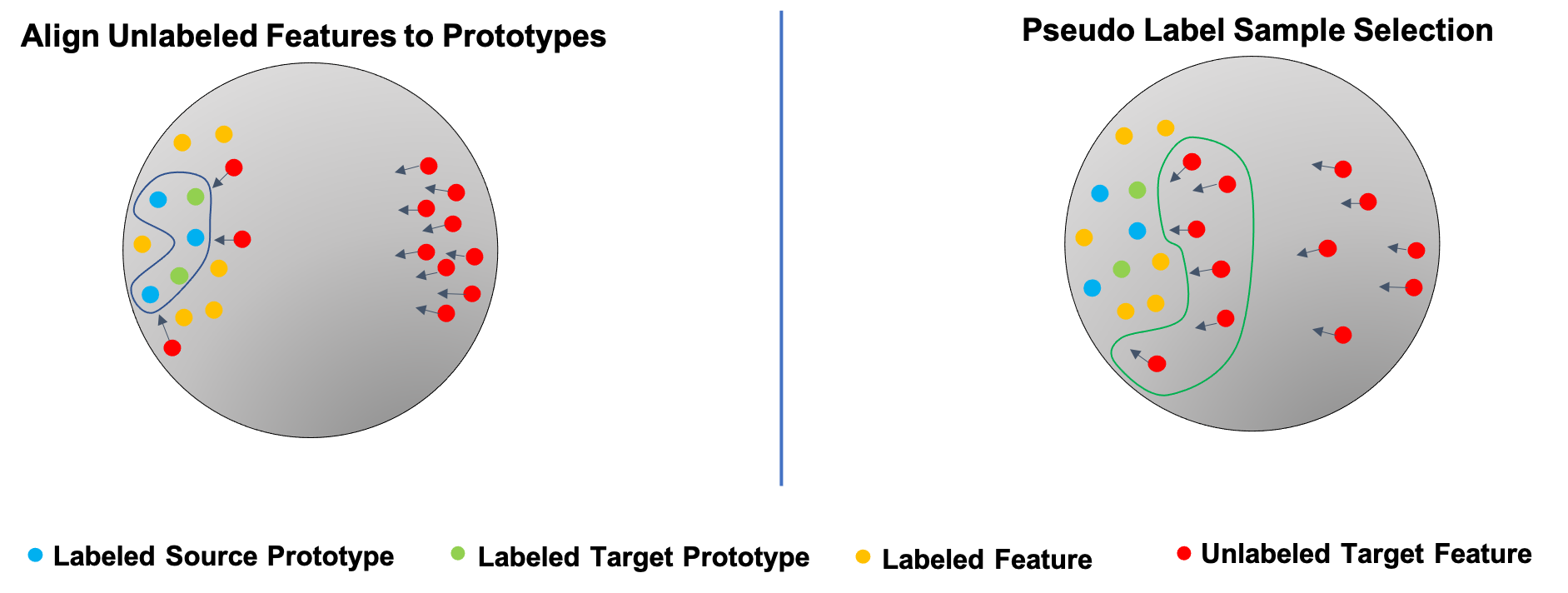}
    \caption{Illustration of Dynamic Feature Alignment. (left) Shows that unlabeled target features (red dots) gradually move to class prototypes of the source (blue dots) and target (green dots). (right) Shows that unlabeled target features with lower entropy values and higher similarity scores (surrounded by the green curve) will be selected for pseudo-label training.}
    \label{fig:framework_detail} 
\end{figure}

\subsection{Dynamic Feature Alignment}
\label{sec:dfa}
Since the labeled target features $f(x^l)$ are already closely aligned to the labeled source features $f(x^s)$~\cite{saito2019semi,kim2020attract}, we focus on the intra-domain discrepancy by directly minimizing the distance between $f(x^u)$ and class prototypes. We choose to use Maximum Mean Discrepancy (MMD)~\cite{gretton2012kernel} to measure the difference between distributions. The basic idea of MMD is that if two distributions are identical, then all the statistics of these two should be the same~\cite{zhu2019aligning}.
\begin{equation}
    D_{\mathcal{H}}(\mathcal{B},\mathcal{D}_{u}) \triangleq \left \| \mathbb{E}_{\mathcal{B}}[\phi(m_{i})] - \mathbb{E}_{\mathcal{D}_{u}}[\phi(f(x_{j}^u))] \right\|_{\mathcal{H}}^{2},
\end{equation}
where $m_{i}$ is the $i$-th feature stored in $\mathcal{B}$, representing the prototype of class $i$, and $f(x_{j}^u)$ is the $j$-th feature in the unlabeled target features $\mathcal{D}_{u}$. Here $\phi (\cdot)$ represents Gaussian Radial Basis Function (RBF) kernels, which map the input feature maps to the reproducing kernel Hilbert space $\mathcal{H}$. The MMD loss can be presented as:
\begin{equation}
    \mathcal{L}_{mmd} = D_{\mathcal{H}}(\mathcal{B},\mathcal{D}_{u}).
\end{equation}
By minimizing $\mathcal{L}_{mmd}$, the domain discrepancy will be bridged and unaligned target features are gradually moving close to the aligned class prototypes. See Fig~\ref{fig:framework_detail} (left) for illustration.

To accelerate the learning process as well as further improve the accuracy, we take advantage of the large number of unlabeled samples and propose a pseudo-label estimation strategy. For an unlabeled image $x^u$, we compute the distances between $f(x^u)$ and every class prototype $m$ stored in $\mathcal{B}$. Here, the cosine similarity is used for distance measurement. Since $m_i$ is the representative prototype of the $i$-th class, a higher similarity value represents a higher probability that $x^u$ belongs to class $i$. Then the pseudo-label estimation function can be formulated by using a softmax function with a temperature $\tau_{p}$ as follows:

\begin{equation}
    P_{dist}(i|x^u) = \frac{exp(m^{T}\cdot f(x^u) / \tau_{p})}{\sum_{j=1}^{K}exp(m^{T}_{j}\cdot f(x^u) / \tau_{p})}.
\end{equation}

Training with inaccurate pseudo-labels can accumulate errors. We adopt a sample selection strategy to keep the high-quality pseudo-labels in the training loop and eliminate the inaccurate ones.  First, those samples with similarity scores higher than the threshold $\epsilon_{dist}$ will be stored in $M_{dist}$. Second, we collect samples with the prediction entropy $H_\mathbf{w}(P_{dist})$ less than a threshold $\epsilon_{ent}$ and store them in $M_{ent}$. This step will selectively collect the samples that are close to the aligned features~\cite{kim2020attract}. See Fig~\ref{fig:framework_detail} (right) for illustration. Last, we take the intersection of $M_{dist}$ and $M_{ent}$, noted as $M_{pse}$. Thus, only samples satisfying both conditions will be used in the pseudo-label training loop. 


The network output $\hat{y}_{x}$ of each sample $x$ in $M_{pse}$ is used as the pseudo-label for calculating the cross entropy loss $\mathcal{L}_{pseudo}$: 

\begin{equation}
    \mathcal{L}_{pseudo} = -\mathbb{E}_{\mathcal{D}_{u}}[\mathbf{1}_{M_{pse}}(x)\log p(y=\hat{y}_{x}|x)].
\end{equation}

To further minimize the intra-domain discrepancy, we follow~\cite{kim2020attract} and apply the same perturbation scheme. We regularize the perturbed features and the raw, clean features by Kullback–Leibler divergence. The goal is to enforce the model to generate perturbation-invariant features so that the perturbation loss can be presented as:
\begin{equation}
    \mathcal{L}_{perturb} = \mathbb{E}_{\mathbf{x}\in\mathcal{D}_{u} \cup\mathcal{D}_{l}}\bigg[\sum_{i=1}^{K}KL[f(x), f(x+r_{x})]\bigg],
\end{equation}
where $x$ is the input image, and $r_x$ is the optimized perturbation added to $x$.
\subsection{Overall Loss Function}
The overall loss function of the proposed DFA framework is the weighted sum of every different piece of loss function mentioned above and can be integrated as follows:

\begin{equation}
 \mathcal{L} = \mathcal{L}_{cls} + \alpha_{1}\mathcal{L}_{mmd} + \alpha_{2}\mathcal{L}_{pseudo} + \alpha_{3}\mathcal{L}_{perturb}.
\end{equation} 

\section{Experiments}
We evaluate our method by conducting experiments on two standard domain adaptation benchmarks. Below we describe the datasets used for these experiments, implementation details, and extensive performance analysis.

\subsection{Datasets}
DomainNet~\cite{peng2019moment} is a large-scale domain adaptation benchmark that contains 6 domains and 345 object categories. Following the previous works MME~\cite{saito2019semi} and APE~\cite{kim2020attract}, we use a 4 domain (Real, Clipart, Painting, Sketch) subset with 126 classes. We report our results on 7 scenarios for a fair comparison with the previous state-of-the-art works. 

We also evaluate our model on Office-Home~\cite{venkateswara2017deep}, which is another domain adaptation benchmark that contains 4 domains (Real, Clipart, Art, Product) and 65 categories. Here we report results on all 12 adaptation scenarios.

\subsection{Implementation Details}
We implement our model using PyTorch~\cite{paszke2019pytorch}. We follow~\cite{saito2019semi, kim2020attract} and report results on DomainNet using AlexNet and ResNet-34 as the backbones. For Office-Home, we report on ResNet-34. All networks are pre-trained on ImageNet. Our model is trained on labeled $\mathcal{D}_{s}$, $\mathcal{D}_{l}$, and unlabeled $\mathcal{D}_{u}$. To make the source and target balanced in the training stage, each mini-batch of labeled samples contains half source samples and half target samples. We consider both one-shot and three-shot setting, and for each class, one (or three) labeled target samples are given for training. For evaluation, we reveal the ground-truth labels of $\mathcal{D}_{u}$ and report results on that. We follow~\cite{saito2019semi, kim2020attract} and use SGD optimizer with an initial learning rate of 0.01, a momentum of 0.9, and a weight decay of 0.0005. For hyperparameters, we set the temperature for the classifier as $\tau=0.05$, set $\gamma=0.1$ to update $\mathcal{B}$ in a fast pace. We set the temperature for pseudo-label estimation as $\tau_p=0.07$. As for thresholds, we set $\epsilon_{dist}$ to $0.3$ for ResNet-34 and $0.1$ for AlexNet, and set $\epsilon_{ent} = 0.5$ for both backbone networks.

\subsection{Baselines}
We compare our proposed method with different models. Baselines include training a network only using labeled samples (S+T), a entropy minimization based semi-supervised method (ENT~\cite{grandvalet2005semi}), three feature alignment-based unsupervised domain adaptation models (DANN~\cite{ganin2015unsupervised}, ADR~\cite{saito2017adversarial}, and CDAN~\cite{long2017conditional}), and three state-of-the-art semi-supervised learning models (SagNet~\cite{nam2021reducing}, MME~\cite{saito2019semi}, and APE~\cite{kim2020attract}) that aim to the same goal as our method.

\begin{table}[!tb]
\setlength{\tabcolsep}{6.5pt}
	\centering
	\caption{Classification accuracy (\%) on the DomainNet dataset for \textbf{three-shot} setting with 4 domains, 7 scenarios using AlexNet and ResNet-34 as backbone networks, respectively.}
	\footnotesize
	\begin{tabular}{c|c|ccccccc|c}
	
    \specialrule{1.25pt}{1pt}{1pt}
    \hline
    
    \textbf{Net} & \textbf{Method} & R to C & R to P & P to C & C to S & S to P & R to S & P to R & MEAN \\ \hline\hline
    \multirow{9}{*}{AlexNet} 
      & S+T & 47.1 & 45.0 & 44.9 & 36.4 & 38.4 & 33.3 & 58.7 & 43.4\\ 
    
    & DANN & 46.1 & 43.8 & 41.0 & 36.5 & 38.9 & 33.4 & 57.3 & 42.4\\ 
    
    & ADR & 46.2 & 44.4 & 43.6 & 36.4 & 38.9 & 32.4 & 57.3 & 42.7\\ 
    
    & CDAN & 46.8 & 45.0 & 42.3 & 29.5 & 33.7 & 31.3 & 58.7 & 41.0\\ 
    
    & ENT & 45.5 & 42.6 & 40.4 & 31.1 & 29.6 & 29.6 & 60.0 & 39.8\\ 
    
    & MME & 55.6 & 49.0 & 51.7 & 39.4 & \textbf{43.0} & 37.9 & 60.7 & 48.2\\ 
    
    & SagNet & 49.1 & 46.7 & 46.3 & 39.4 & 39.8 & 37.5 & 57.0 & 45.1\\ 
    
    & APE & 54.6 & 50.5 & \textbf{52.1} & 42.6 & 42.2 & 38.7 & 61.4 & 48.9\\ 
    
    & Ours & \textbf{55.0} & \textbf{52.3} & 51.6 & \textbf{44.5} & 41.8 & \textbf{39.4} & \textbf{62.1} & \textbf{49.5}\\ \hline\hline

    \multirow{9}{*}{ResNet} 
      & S+T & 60.0 & 62.2 & 59.4 & 55.0 & 59.5 & 50.1 & 73.9 & 60.0\\ 
    
    & DANN & 59.8 & 62.8 & 59.6 & 55.4 & 59.9 & 54.9 & 72.2 & 60.7\\ 
    
    & ADR & 60.7 & 61.9 & 60.7 & 54.4 & 59.9 & 51.1 & 74.2 & 60.4\\ 
    
    & CDAN & 69.0 & 67.3 & 68.4 & 57.8 & 65.3 & 59.0 & 78.5 & 66.5\\ 
    
    & ENT & 71.0 & 69.2 & 71.1 & 60.0 & 62.1 & 61.1 & 78.6 & 67.6\\ 
    
    & MME & 72.2 & 69.7 & 71.7 & 61.8 & 66.8 & 61.9 & 78.5 & 68.9\\ 
    
    & SagNet & 62.0 & 62.9 & 61.5 & 57.1 & 59.0 & 54.4 & 73.4 & 61.5\\ 
    
    & APE & 76.6 & 72.1 & \textbf{76.7} & 63.1 & 66.1 & \textbf{67.5} & 79.4 & 71.7\\ 
    
    & Ours & \textbf{76.7} & \textbf{73.9}& 75.4 & \textbf{65.5} & \textbf{70.5} & \textbf{67.5} & \textbf{80.3} & \textbf{72.8} \\ \hline 

    \specialrule{1.25pt}{1pt}{1pt}
 
  \end{tabular}

  \label{table:domainnet}
\end{table}

\begin{table}[!tb]
\setlength{\tabcolsep}{6.5pt}
	\centering
	\caption{Classification accuracy (\%) on the DomainNet dataset for \textbf{one-shot} setting with 4 domains, 7 scenarios using ResNet-34.}
	\footnotesize
	\begin{tabular}{c|c|ccccccc|c}
	
    \specialrule{1.25pt}{1pt}{1pt}
    \hline
    
    \textbf{Net} & \textbf{Method} & R to C & R to P & P to C & C to S & S to P & R to S & P to R & MEAN \\ \hline\hline

    \multirow{9}{*}{ResNet} 
      & S+T & 55.6 & 60.6 & 56.8 & 50.8 & 56.0 & 46.3 & 71.8 & 56.9\\ 
    
    & DANN & 58.2 & 61.4 & 56.3 & 52.8 & 57.4 & 52.2 & 70.3 & 58.4\\ 
    
    & ADR & 57.1 & 61.3 & 57.0 & 51.0 & 56.0 & 49.0 & 72.0 & 57.6\\ 
    
    & CDAN & 65.0 & 64.9 & 63.7 & 53.1 & 63.4 & 54.5 & 73.2 & 62.5\\ 
    
    & ENT & 65.2 & 65.9 & 65.4 & 54.6 & 59.7 & 52.1 & 75.0 & 62.6\\ 
    
    & MME & 70.0 & 67.7 & 69.0 & 56.3 & 64.8 & 61.0 & 76.1 & 66.4\\ 
    
    & SagNet & 59.4 & 61.9 & 59.1 & 54.0 & 56.6 & 49.7 & 72.2 & 59.0\\ 
    
    & APE & 70.4 & 70.8 & \textbf{72.9} & 56.7 & 64.5 & \textbf{63.0} & 76.6 & 67.6\\ 
    
    & Ours & \textbf{71.8} & \textbf{72.7} & 69.8 & \textbf{60.8} & \textbf{68.0} & 62.3 & \textbf{76.8} & \textbf{68.9} \\ \hline 

    \specialrule{1.25pt}{1pt}{1pt}
 
  \end{tabular}

  \label{table:domainnet-1shot}
\end{table}

\begin{table}[]
	\caption{Classification accuracy (\%) comparisons on Office-Home for \textbf{three-shot} setting with 4 domains, 12 scenarios using ResNet-34 as the backbone network.}
\setlength{\tabcolsep}{2pt}
\footnotesize	
\centering
\begin{tabular}{c|cccccccccccc|c}
\specialrule{1.25pt}{1pt}{1pt}
\hline
\textbf{Method} & R to C        & R to P        & R to A        & P to R        & P to C        & P to A        & A to P        & A to C        & A to R        & C to R        & C to A        & C to P        & MEAN \\ \hline\hline
S+T             & 55.7          & 80.8          & 67.8          & 73.1          & 53.8          & 63.5          & 73.1          & 54.0          & 74.2          & 68.3          & 57.6          & 72.3          & 66.2 \\
DANN            & 57.3          & 75.5          & 65.2          & 69.2          & 51.8          & 56.6          & 68.3          & 54.7          & 73.8          & 67.1          & 55.1          & 67.5          & 63.5 \\
CDAN            & 61.4          & 80.7          & 67.1          & 76.8          & 58.1          & 61.4          & 74.1          & 59.2          & 74.1          & 70.7          & 60.5          & 74.5          & 68.2 \\
ENT             & 62.6          & 85.7          & 70.2          & 79.9          & 60.5          & 63.9          & 79.5          & 62.3          & 79.1          & 76.4          & 64.7          & 79.1          & 71.9 \\
MME             & 64.6          & 85.5          & 71.3          & 80.1          & 64.6          & 65.5          & 79.0          & 63.6          & 79.7          & 76.6          & 67.2          & 79.3          & 73.1 \\
APE             & 66.4          & 86.2          & 73.4          & 82.0          & 65.2          & 66.1          & \textbf{81.1} & \textbf{63.9}          & 80.2 & \textbf{76.8} & 66.6          & \textbf{79.9} & 74.0 \\
Ours            & \textbf{68.3} & \textbf{86.9} & \textbf{74.1} & \textbf{82.3} & \textbf{65.9} & \textbf{67.8} & 80.4          & 63.0 & \textbf{80.3}         & 76.6          & \textbf{67.8} & 79.1         &   \textbf{74.4}   \\ 
\hline
\specialrule{1.25pt}{1pt}{1pt}

\end{tabular}
\label{table:officehome}
\end{table}
\subsection{Experiment Results}

We summarize the comparisons between our method and the baselines on 7 adaptation scenarios of the DomainNet dataset in Table~\ref{table:domainnet} and Table~\ref{table:domainnet-1shot}, for three-shot setting and one-shot setting respectively. When using the ResNet-34 as the backbone network, our method outperforms the current state-of-the-art baseline by more than $1\%$, and achieves the best performance in most adaptation scenarios. For the best case \textit{S to P} of three-shot setting, our method surpasses the second-best method by $4.4\%$. When AlexNet is applied as the backbone network, the margin that our method outperforms other methods is not as large as using ResNet-34 because our proposed scheme requires high-quality intermediate features stored in $\mathcal{B}$, and ResNet-34 has more advantage in achieving that compared with AlexNet. Our method still performs the best in most scenarios, surpassing APE by $0.6\%$ on average. 

The comparison results of our method with other baselines of 12 adaptation scenarios on Office-Home dataset are summarized in Table~\ref{table:officehome}. We report results using ResNet-34 as the backbone network. Our method achieves the best performance on 8 out of 12 adaptation scenarios and outperforms all the baselines on average.

\subsection{Analysis}
In Sec~\ref{sec: bank}, we explain the updating rules of the dynamic memory bank $\mathcal{B}$. Here we conduct an experiment to show how $\gamma$ affects the classification accuracy. We use AlexNet as the backbone network and train the model on DomainNet with different $\gamma$. A smaller $\gamma$ means updating $\mathcal{B}$ faster, and larger $\gamma$ means updating $\mathcal{B}$ in a more stable way. We evaluate on 3 adaptation scenarios and summarize the results in Table~\ref{table:pace}. It demonstrates that a more stable $\mathcal{B}$ (with larger $\gamma$ as 0.75) can not help improve the performance, and replacing the entire $\mathcal{B}$ with the new one every iteration ($\gamma=0$) can not get the optimal performance either. Our results show that using a small (but larger than $0$) $\gamma$ and updating $\mathcal{B}$ in a relatively faster pace achieves the best classification accuracy.

\begin{table}[]
	\caption{Analysis on how $\gamma$ affects the classification accuracy ($\%$).}
\setlength{\tabcolsep}{6pt}
\footnotesize	
\centering
\begin{tabular}{c|ccc|c}
\specialrule{1.25pt}{1pt}{1pt}
\hline
\textbf{Method} & R to C        & R to P        & C to S        & MEAN \\ \hline\hline
APE             & 54.6          & 50.5          & 42.6          & 49.2 \\
Ours ($\gamma=0.75$)             & 54.1          & 50.7          & 42.4          & 49.0 \\
Ours ($\gamma=0.25$)             & 54.9          & 52.2          & 43.6          & 50.2 \\
Ours ($\gamma=0.1$)            & \textbf{55.0} & \textbf{52.3} & \textbf{44.5} & \textbf{50.6} \\ 
Ours ($\gamma=0$)             & 54.7          & 52.0          & \textbf{44.5}          & 50.4 \\
\hline
\specialrule{1.25pt}{1pt}{1pt}

\end{tabular}
\label{table:pace}
\end{table}

\begin{figure*}
    \centering
    \begin{tabular}{cc}
        \includegraphics[width=0.4\linewidth]{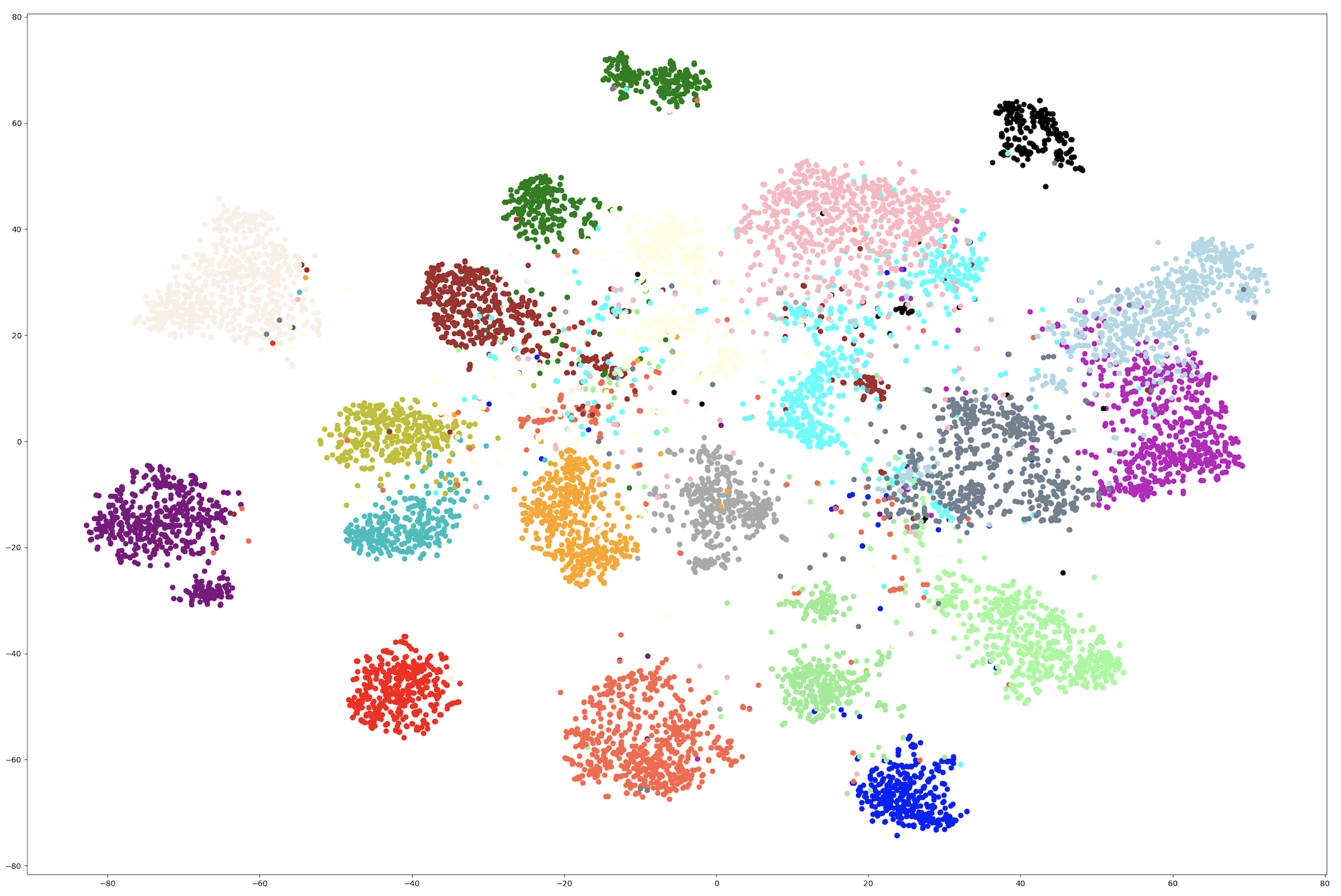} & 
        \includegraphics[width=0.4\linewidth]{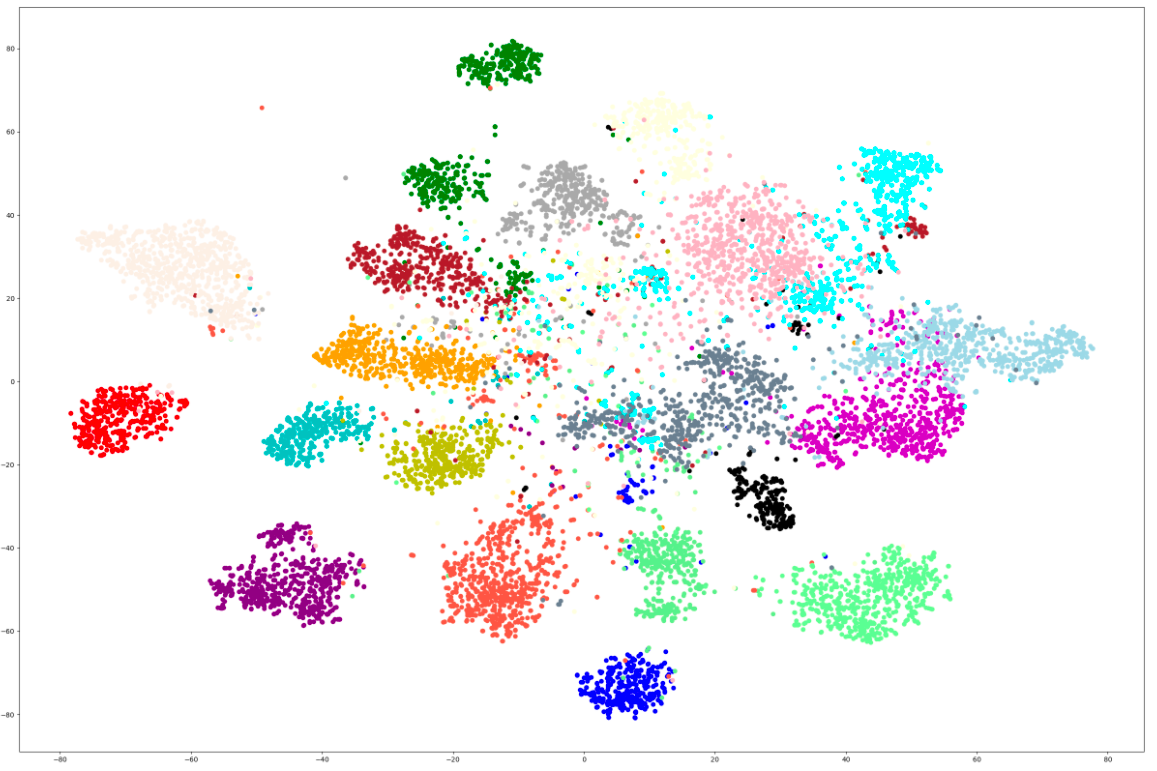} \\
        \small{Epoch 1} & \small{Epoch 10}
        \\
        \includegraphics[width=0.4\linewidth]{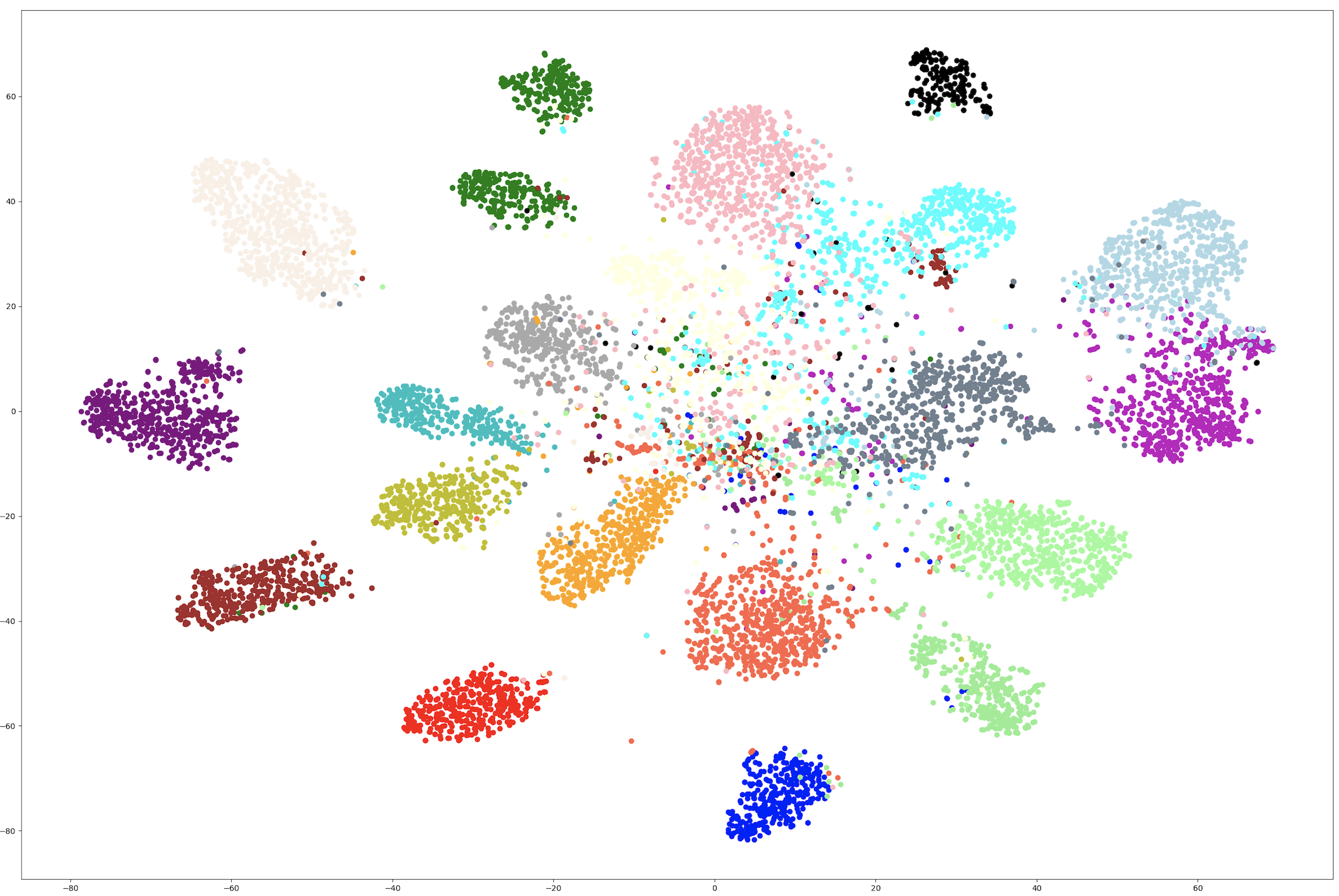}
        &
        \includegraphics[width=0.4\linewidth]{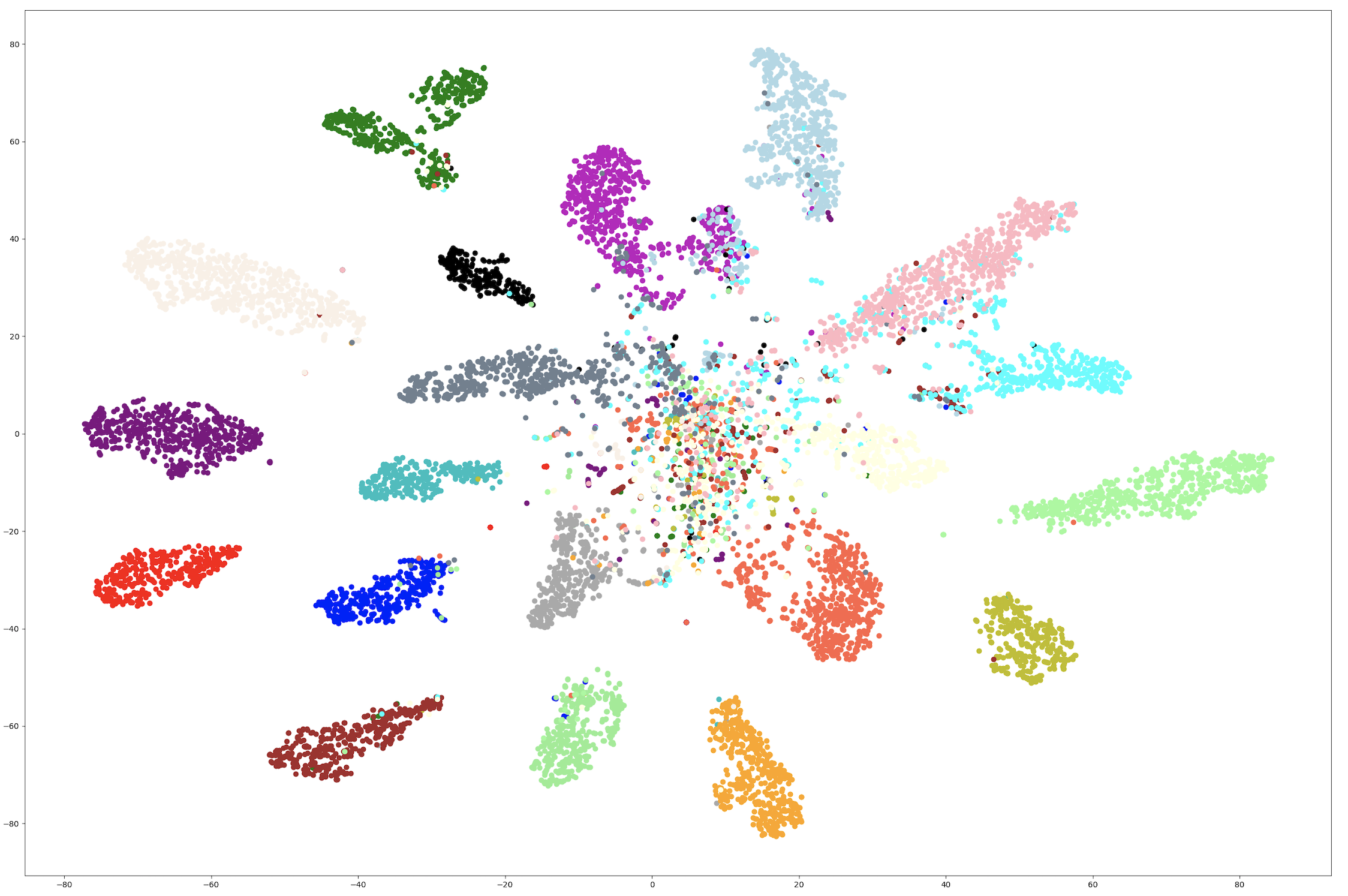}\\
        \small{Epoch 50} & \small{Epoch 100}
    \end{tabular}
    \caption{The t-SNE visualization of intermediate features in the target domain of our method at different training stages.}
    \label{fig:tsne}
\end{figure*}

To better understand the feature alignment progress, we show the t-SNE~\cite{van2008visualizing} embedding of the intermediate features at different training stages in Fig~\ref{fig:tsne}. We visualize the target features extracted by ResNet-34 in the experiment of \textit{Painting to Real} scenario of the DomainNet. Following APE~\cite{kim2020attract}, we randomly select 20 classes out of 126 classes in the dataset for clarity. This shows that as training progresses, the target feature clusters will gradually be split for better classification.

\section{Conclusions}

We proposed a novel approach for semi-supervised domain adaptation that uses a dynamic memory bank to support inter- and intra-domain feature alignment. Our update approach is designed to be class balanced, thereby mitigating one of the more challenging aspects of the problem. We evaluated our approach on two standard datasets and found that it significantly improved the average accuracy over the previous state-of-the-art techniques. In addition to improved accuracy, our approach has several attractive features. It doesn't require significant additional memory (only two copies of the class prototypes) or computation (only online updates to the intermediate and dynamic memory banks). It also doesn't require extensive parameter tuning: the weights for the loss function are fixed across all experiments, accuracy isn't particularly sensitive to the dynamic memory updates parameter $\gamma$, and the pseudo-label thresholds only needed to be adjusted to account for the low discriminative power of AlexNet. Given this, we believe this approach will be applicable to many semi-supervised domain adaptation scenarios.



\subsection*{Acknowledgements}

This material is based upon work supported by the National Science Foundation under Grant No.\ IIS-1553116.

\bibliography{citation.bib}
\end{document}